\documentclass{article}

\usepackage{PRIMEarxiv}
\usepackage{times}  % DO NOT CHANGE THIS
\usepackage{amssymb}
\usepackage{amsmath}
\usepackage{booktabs}
\usepackage{multirow}
\usepackage[utf8]{inputenc} % allow utf-8 input
\usepackage[T1]{fontenc}    % use 8-bit T1 fonts
\usepackage{hyperref}       % hyperlinks
\usepackage{url}            % simple URL typesetting
\usepackage{booktabs}       % professional-quality tables
\usepackage{amsfonts}       % blackboard math symbols
\usepackage{nicefrac}       % compact symbols for 1/2, etc.
\usepackage{microtype}      % microtypography
\usepackage{lipsum}
\usepackage{fancyhdr}       % header
\usepackage[table]{xcolor}
\usepackage{pifont}
\usepackage[most]{tcolorbox}
\usepackage{graphicx}       % graphics
\graphicspath{{media/}}     % organize your images and other figures under media/ folder

\title{Towards Training-free Multimodal Hate Localisation with Large Language Models
\thanks{\textit{\underline{Citation}}: 
\textbf{Sun et al. Towards Training-free Multimodal Hate Detection with Large Language Models}} 
}

\author{
  Yueming Sun \\
  Hybrid Intelligence Lab, University of Durham \\
  Multimodal Intelligence Lab, University of Exeter \\
  \texttt{yueming.sun@durham.ac.uk} \\
  \And
  Long Yang \\
  Hybrid Intelligence Lab \\
  University of Durham \\
  \texttt{yang.long@durham.ac.uk} \\
  \And
  Jianbo Jiao \\
  The MIx Group \\
  University of Birmingham \\
  \texttt{j.jiao@bham.ac.uk} \\
  \And
  Zeyu Fu \\
  Multimodal Intelligence Lab \\
  University of Exeter \\
  \texttt{z.fu@exeter.ac.uk} \\
}

\begin{document}
\maketitle

\begin{abstract}
The proliferation of hateful content in online videos poses severe threats to individual well-being and societal harmony. However, existing solutions for video hate detection either rely heavily on large-scale human annotations or lack fine-grained temporal precision. In this work, we propose \textbf{LELA}, the first training-free \textbf{L}arge Language Model (LLM) based framework for hat\textbf{e} video \textbf{l}ocaliz\textbf{a}tion.  Distinct from state-of-the-art models that depend on supervised pipelines, LELA leverages LLMs and modality-specific captioning to detect and temporally localize hateful content in a training-free manner. Our method decomposes a video into five modalities, including image, speech, OCR, music, and video context, and uses a multi-stage prompting scheme to compute fine-grained hateful scores for each frame. We further introduce a composition matching mechanism to enhance cross-modal reasoning. Experiments on two challenging benchmarks, HateMM and MultiHateClip, demonstrate that LELA outperforms all existing training-free baselines by a large margin.
% achieving up to 72.64\% ROC-AUC. 
We also provide extensive ablations and qualitative visualizations, establishing LELA as a strong foundation for scalable and interpretable hate video localization.

\textcolor{red}{\textbf{Disclaimer:} This paper discusses hateful or offensive content that may be disturbing to some readers. Viewer discretion is advised.}
\end{abstract}

\section{Introduction}
\begin{figure}
    \centering
    \includegraphics[width=\linewidth]{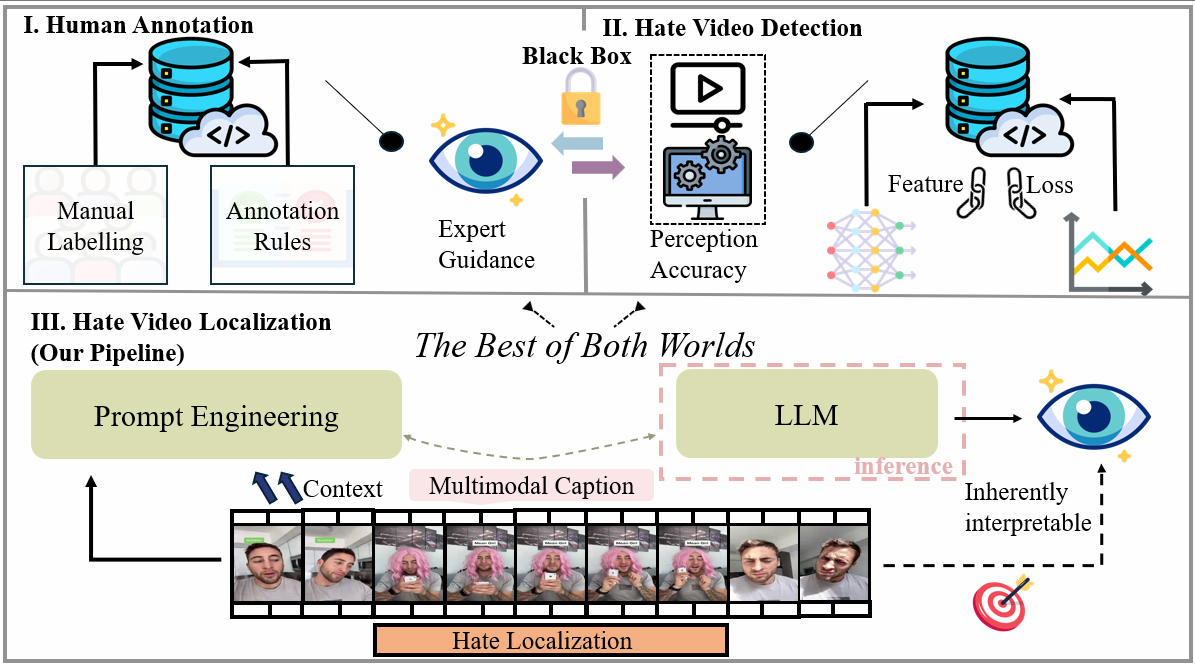}
    \caption{\small We propose the first LLM-based framework for video hate localization, which addresses the challenge of interpretability in hate content moderation and locates frame-level hateful content from multimodal video input.}
    \label{fig:advert}
\end{figure}
While significant efforts have been devoted to detecting textual hate speech~\cite{waseem2016racist,sharma2022harmfulmemes}, hateful video detection remains comparatively underexplored. Unlike text, videos contain multiple synchronized modalities, including visuals, speech, music, and on-screen text, which together can convey hate in implicit, context-dependent ways~\cite{singh2022transformerviolence}. Manual moderation of such content is time-consuming, psychologically harmful, and ultimately non-scalable, motivating automatic multimodal hate detection frameworks for safer digital environments.

 Recent benchmarks such as HateMM~\cite{hatemm} and MultiHateClip~\cite{multihateclip} mark important progress by providing snippet-level annotations across multiple modalities. HateMM includes videos with visual, auditory, and textual elements, while MultiHateClip comprises short videos annotated for hatefulness, offensive language, victim targeting, and modality attribution in both English and Chinese. However, existing methods~\cite{multihateclip,hatemm,lang2025biting,hee2025contrastive} trained on these datasets still rely predominantly on video-level or snippet-level labels, overlooking richer temporal granularity. As shown in Figure 1, such video-level classification systems essentially act as black boxes: the model only outputs a label, while it remains unclear why or where in the video this decision is made, casting doubt on whether the model’s reasoning truly aligns with human understanding. This motivates the exploration of video hate localization, which explicitly targets frame-level understanding of hateful content, which targets frame-level understanding of hateful content. Our approach adopts a training-free paradigm based on LLMs to produce frame-wise predictions, thereby providing greater explainability of model decisions and enabling a more comprehensive evaluation of hateful content understanding beyond coarse video-level accuracy.

\begin{figure*}
    \centering
    \includegraphics[width=0.95\linewidth]{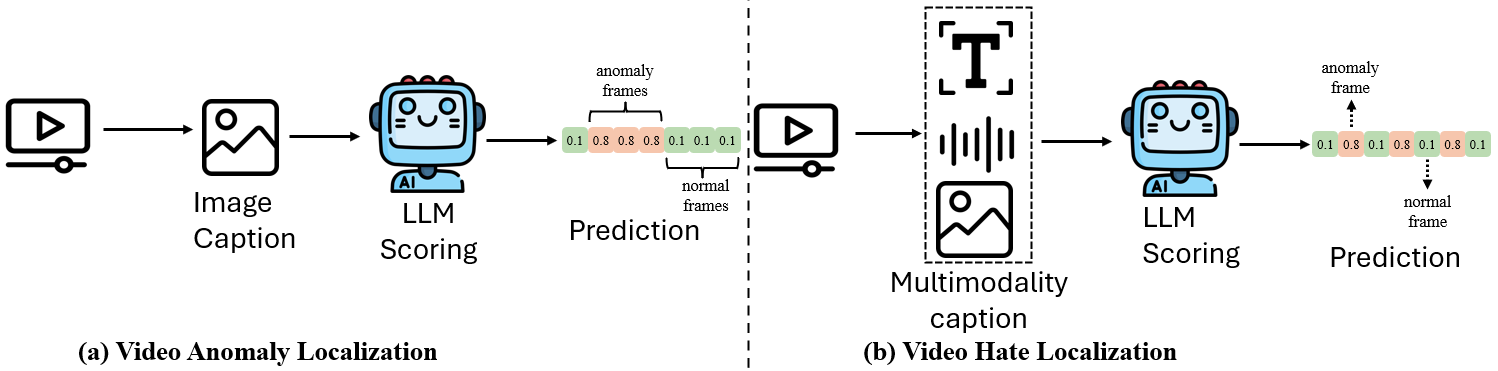}
    \caption{Conventional video anomaly detection (a) focuses on visual signals and aims to detect short abnormal snippets within an otherwise normal video, typically producing coarse video-level anomaly scores. Video hate localization (b) jointly leverages multiple modalities (image, audio, and text), attends to high-level semantic cues related to hateful or offensive content, and assigns fine-grained frame-level scores, enabling precise localization of harmful regions within the video.}
    \label{fig:pipeline}
\end{figure*}

Unlike video anomaly detection (VAD) , which is an earlier task with many strong baselines \cite{LAVAD,VERA,lin2025unified}, video malicious content detection is inherently more semantic and context-sensitive. Conventional VAD mainly focuses on visual (and occasionally textual) patterns and analyzes whether certain frames deviate from the overall video distribution. In contrast, video malicious content detection relies on coordinated multimodal signals and requires segment-level judgments informed by real-world knowledge of harmful or offensive behavior. As illustrated in Fig.~\ref{fig:pipeline}, VAD primarily aims to identify short unusual snippets within otherwise normal videos, whereas video hate localization operates over multiple modalities, attends to high-level cues related to hateful content, and independently ranks or labels each frame.

In summary, in this paper we introduce the task of hateful video localization, 
 where the goal is not only to predict whether a video is hateful at the video level, but also to assign interpretable hate scores to individual frames or short segments. Given a multimodal video composed of visual frames, audio, and automatically generated captions, the model must produce a temporally resolved profile of hatefulness that highlights the precise regions responsible for the final decision. Such frame-level or snippet-level localization naturally supports human-in-the-loop review, evidence highlighting in appeal procedures, and model auditing, and can also serve as a fine-grained supervision signal for future supervised models.

Building on these insights, our contributions in this paper are threefold:
\begin{itemize}
    \item We introduce the first LLM-based framework for temporal localization of hateful content in videos, eliminating the reliance on supervised annotations and task-specific model tuning, and making it highly adaptable to real-world scenarios.
    \item We propose a multi-stage prompting mechanism and a composition matching strategy that exploit modality-specific captions and harness the reasoning capabilities of LLMs to perform fine-grained hate scoring with improved interpretability.
    \item We conduct extensive experiments on the HateMM and MHC-English datasets, delivering the first comprehensive evaluation of LLMs’ ability to detect and localize hateful content across multiple modalities, languages, and cultural contexts.
\end{itemize}

\section{Related Work}
\subsection{Multimodal Hate Video Detection}
Hate speech detection, traditionally dominated by text-based methods~\cite{badjatiya2017deep}, has increasingly shifted towards multimodal scenarios due to the proliferation of harmful visual content, such as memes and videos~\cite{kiela2020hateful, gomez2020exploring}. Recent efforts have introduced comprehensive datasets, such as Facebook's Hateful Memes (FHM)~\cite{kiela2020hateful} and the Multimedia Automatic Misogyny Identification (MAMI) dataset~\cite{fersini2022semeval}, enabling the development of robust multimodal models that jointly reason over images and text. These benchmarks have revealed that hateful intent is often expressed through subtle interactions between visual symbolism and linguistic cues, rather than through explicit slurs alone.

Building on these insights, subsequent work has begun to extend multimodal hate detection from static images to dynamic video content. Wang et al.~\cite{wang2025crossmodal} addressed data scarcity issues by leveraging meme datasets to augment limited hateful video data, demonstrating substantial performance gains through cross-modal transfer learning and highlighting the strong inductive bias shared between short-form memes and video frames. The MultiHateClip (MHC)~\cite{lee2024multihateclip}, and HateMM~\cite{sabatelli2024hatemm} datasets provided valuable benchmarks for this line of research by introducing multimodal video annotations, although their relatively small size and domain-specific sources still present significant challenges for generalization. Advances in multimodal fusion methods~\cite{muennighoff2022vlm, li2023videochat} and large-scale vision-language pre-training~\cite{radford2021learning, li2022blip} further enhance multimodal hate speech detection by enabling better integration of visual and textual signals through shared embedding spaces and cross-attention mechanisms.

However, most existing methods remain confined to video-level classification and fail to capture the temporal dynamics of hateful expressions, which often emerge only in specific moments or in the interplay between a sequence of frames and spoken content. Furthermore, the reliance on high-level embeddings from frozen vision-language models often leads to shallow cross-modal alignment, limiting the model’s ability to understand subtle, context-dependent hate signals or to provide interpretable justifications for its predictions. In contrast, our work explicitly targets frame-level localization of hateful content and leverages LLMs to reason over rich, modality-specific captions, thereby moving beyond coarse video-level labels towards temporally fine-grained and more interpretable judgments.

\subsection{Training-Free Video Anomaly Detection}
Traditional video anomaly detection (VAD) methods usually require extensive data labeling and model training, often in the form of one-class classifiers, reconstruction-based models, or predictive networks that learn a notion of normality from large collections of benign videos. While effective in constrained settings, these approaches are typically domain-specific and struggle to generalize to new environments or anomaly types without costly retraining. Recently, training-free approaches leveraging large-scale pre-trained models have emerged as an attractive alternative, as they can reuse powerful generic representations acquired from diverse web-scale data.

LAVAD~\cite{LAVAD} proposed a training-free method that utilizes vision-language models (VLMs) to caption frames, followed by anomaly score estimation through large language models (LLMs). Their method significantly outperformed traditional unsupervised and one-class classification methods on standard benchmarks, illustrating the feasibility of VAD without explicit task-specific training. Similar training-free paradigms were explored by leveraging pre-trained VLMs and LLMs to describe and analyze video content effectively~\cite{menapace2023, tur2023diffusion, tur2023reconstruction}, for example by comparing predicted descriptions against normality priors, or by using generative models to measure reconstruction difficulty as a surrogate for abnormality.

Despite these advances, training-free approaches are still constrained by the quality of generated descriptions and the cross-modal alignment capabilities of the underlying models, which may miss domain-specific cues or subtle semantic anomalies. Moreover, while VAD typically focuses on detecting deviations from expected patterns—often relying on low-level visual or motion cues—hateful content detection demands high-level semantic reasoning across multiple modalities and sensitivity to social, cultural, and legal norms. The subjective, context-dependent nature of hate further complicates this task, making the straightforward application of training-free VAD frameworks insufficient for accurate and interpretable hate detection. Our work can be viewed as extending the training-free paradigm from generic anomaly detection to the more challenging setting of hate localization, with dedicated prompting and aggregation strategies tailored to this safety-critical domain.

\subsection{LLMs for Video Understanding}
Large language models (LLMs) have revolutionized natural language processing tasks, demonstrating strong reasoning, abstraction, and generalization capabilities~\cite{brown2020language}. Recent research increasingly explores integrating LLMs with visual data to advance video understanding tasks, such as anomaly detection, captioning, and multimodal content moderation~\cite{LAVAD, wang2025crossmodal}. Menapace et al.~\cite{menapace2023} utilized LLMs to aggregate temporal information and generate contextually rich descriptions for anomaly detection, showing that textual summarization of visual content can serve as an effective intermediate representation for downstream reasoning. Likewise, Lee et al.~\cite{lee2024multihateclip} leveraged LLMs and vision-language pre-training to enhance hateful content identification in videos, indicating that LLMs can capture nuanced semantic relationships across modalities when provided with appropriate prompts.

Beyond these specific applications, a broader line of work investigates multimodal LLMs and video-centric architectures~\cite{li2023videochat, muennighoff2022vlm, alayrac2022flamingo}, where video frames are encoded into visual tokens and fed into LLM backbones to enable open-ended question answering, dialogue, and reasoning about dynamic scenes. These approaches illustrate the potential of LLMs not only for content analysis but also for their adaptability to diverse downstream tasks, driven by pre-training on large-scale multimodal corpora and instruction tuning. Nevertheless, challenges remain, particularly regarding the effective integration of temporal information, the handling of long videos, and the calibration of LLM outputs for safety-sensitive tasks~\cite{wang2025crossmodal, llava-1.5}. In this work, we build on these developments by designing a multi-stage prompting framework that decomposes hate localization into interpretable sub-tasks, and by combining LLM-based scoring with modality-aware composition matching to better exploit temporal and multimodal cues.

\begin{figure*}[t]
  \includegraphics[width=\textwidth]{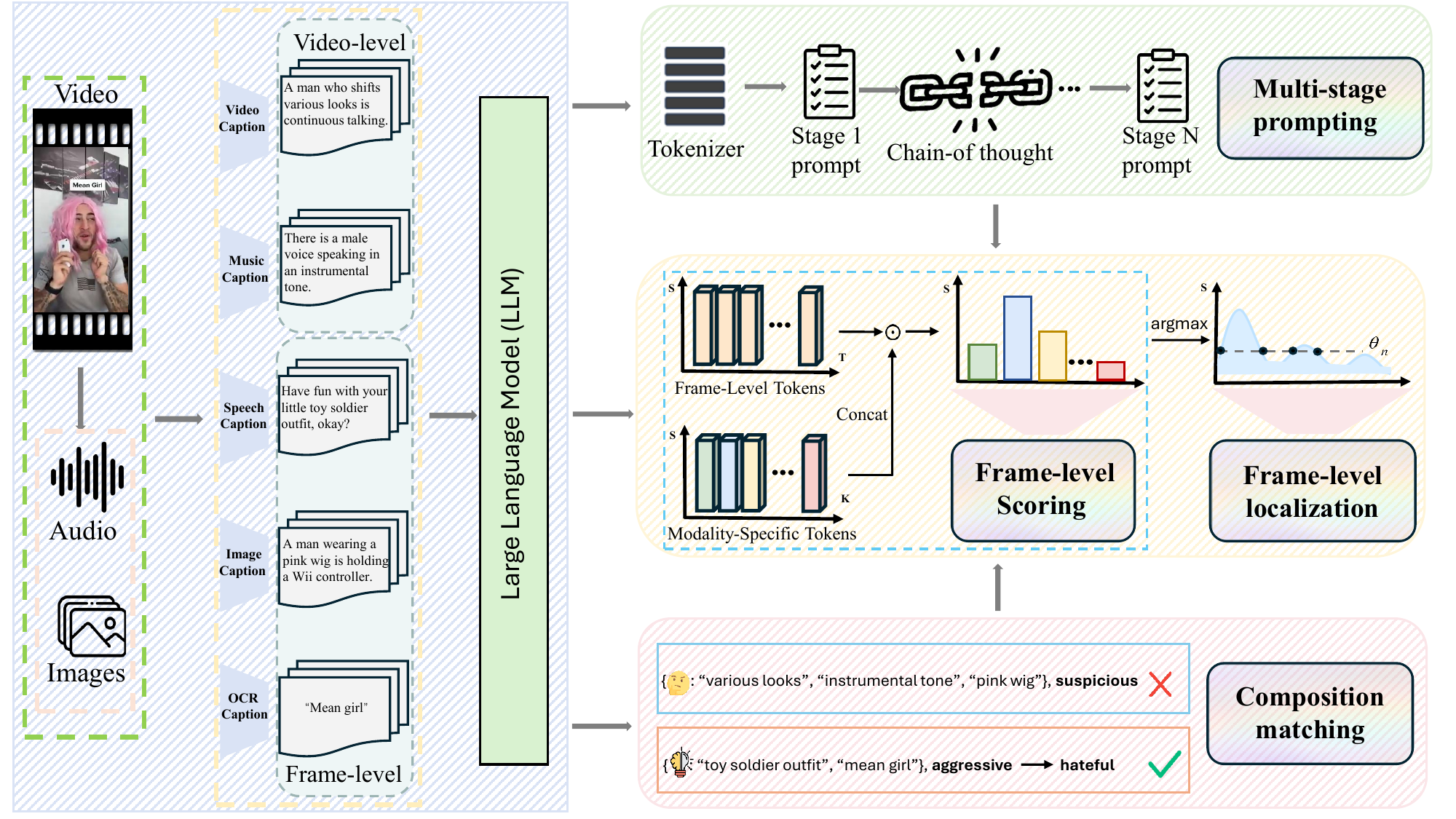}
  \caption{\small Overview of the LELA framework. Given an input video, LELA decomposes it into five modalities: video snippets, static images, background music, speech, and OCR text. Each modality is processed by a dedicated captioning model to extract frame-aligned textual descriptions. A multi-stage prompting strategy (green) evaluates frame-level modality captions from complementary perspectives (e.g., explicit hate, implied hate, target groups), while a parallel composition matching module (pink) summarizes salient information across time and modalities. The resulting textual representations are fed into an LLM to produce per-modality scores, and the final frame-level hate score is obtained by taking the maximum across modalities, enabling fine-grained localization of hateful content and providing interpretable evidence for moderation decisions.}
  \label{fig:architecture}
\end{figure*}
\section{Methodology}

\subsection{Problem Formulation}

Given an input video \( V \), which comprises multiple modalities, including visual frames \( I = [I_1, I_2, \dots, I_M] \in V \), textual information \( T = [T_1, T_2, \dots, T_M] \in V \), and audio segments \( A = [A_1, A_2, \dots, A_M] \in V \), where \( M \) denotes the total number of frames, existing methods for detecting hateful content seek to learn a classification function \( f \) capable of assigning each modality-specific input frame to either the normal (non-hateful, score 0) or hateful (score 1) category. Typically, the detection model \( f \) is trained using a labeled dataset \(\mathcal{D}\) consisting of pairs of the form \((V_i, y_i)\), where \( V_i \) represents a video sequence and \( y_i \in \{0,1\} \) denotes its corresponding binary hate label. 

Formally, given a dataset: $\mathcal{D} = \{(V_i, y_i)\}_{i=1}^{N}$, the training process aims to identify the optimal function \( f^* \) that accurately maps video inputs to their associated labels by minimizing a loss function \(\mathcal{L}\):
\begin{equation}
f^* = \arg\min_{f \in \mathcal{F}} \mathcal{L}(f(V), y)
\end{equation}
where \(\mathcal{F}\) denotes the hypothesis space of possible classification functions and \(\mathcal{L}\) is the loss function. 
The granularity of annotation labels \( y \) in existing datasets can vary significantly based on the supervision strategy. Labels may be provided as dense binary vectors indicating frame-level annotations or as coarse, single binary labels corresponding to entire video clips. However, current approaches predominantly rely on video-level annotations due to the considerable computational overhead involved in generating fine-grained, frame-level predictions across entire video sequences. 
% Furthermore, the subjective nature inherent to human annotation processes introduces variability and inconsistency into the labeling, necessitating the creation of extensive, carefully annotated datasets to mitigate this noise and facilitate effective model training. Nevertheless, constructing such datasets remains challenging due to the scarcity and rare occurrence of explicitly hateful content, resulting in resource-intensive labeling processes. 
% Additionally, ethical considerations and privacy constraints further complicate the collection, annotation, and dissemination of video data (\(V\)), posing significant obstacles to developing comprehensive, large-scale datasets suitable for training robust hate-detection models.

In contrast to conventional methods, we introduce a novel framework for hate video localiszation, referred to as \textit{training-free Hate Video Locaslization (HVL)}. This paradigm aims to estimate a hate-related score for each frame $I \in V$ by utilizing pre-trained models solely at inference time, \textit{i.e.}, without requiring any training or fine-tuning on a labeled dataset $\mathcal{D}$.

\subsection{Composition Matching}

Previous empirical experimentation \cite{botelho2021deciphering} has observed that multimodal components play a crucial role in hate detection. Some videos may contain no obvious offensive content in their text or visuals individually, yet their combination can give rise to new meanings. While previous fusion-based studies \cite{chhabra2024mhs} \cite{van2025detecting} have primarily considered general visual and audio modalities, there remains limited understanding regarding how encoded embeddings concentrate specifically on detailed textual content. To address this gap and emphasize the explicit textual information within multimedia, we propose decomposing the conventional vision and audio modalities into more fine-grained textual captions.

Specifically, we subdivide the visual modality into image captions and OCR (optical character recognition) captions, and the audio modality into speech captions and music captions. For visual modalities, we employ BLIP-2 ~\cite{blip-2} to generate comprehensive image captions that provide general descriptions of visual scenes, while using EasyOCR ~\cite{easyocr} as our OCR captioning model to accurately extract text appearing in visual content. Regarding audio modalities, we apply the Whisper~\cite{Whisper} model to transcribe spoken language (speech captions) and leverage LP-Music-Caps~\cite{LP-music} for generating descriptive captions of background music and environmental sounds. Given that individual frames represent static visual snapshots, we utilize the PDVC~\cite{PDVC} model as a dedicated video captioning approach to describe dynamic events across video frames. By aligning these diverse captioning outputs across multiple modalities, our proposed composition matching approach enhances the representation of explicit textual signals within the multimodal embedding space, significantly improving the interpretability and accuracy of hateful content detection.

To enhance explicit textual alignment across modalities, we concatenate modality-specific captions with speech captions at each frame \(j\):
\begin{equation}
    \hat{C}_j^{m} = \text{Concat}(C_j^{\text{speech}}, C_j^{m}), \quad m \in \{\text{img}, \text{ocr}, \text{music}, \text{video}\}.
    \label{eq:concat_caption}
\end{equation}
Subsequently, we summarize these concatenated captions using LLM (\(\Phi_{\text{LLM}}^{\text{sum}}\)):
\begin{equation}
    C_j^{m, \text{sum}} = \Phi_{\text{LLM}}^{\text{sum}}(\hat{C}_j^{m})
    \label{eq:summary}
\end{equation}
producing concise yet rich semantic representations for subsequent scoring via multi-stage prompting.

\subsection{Multi-stage Prompting}
The primary objective of our task is to accurately quantify the degree of hateful content within multimodal contexts by assigning a continuous score ranging from 0 (non-hateful) to 1 (hateful). Based on preliminary study (see Appendix),  we observe that conventional single-stage prompting exhibits limited performance in hate detection tasks. To address this issue, we propose decomposing the reasoning process into multiple stages, enabling more fine-grained and interpretable evaluation. Inspired by the chain-of-thought prompting framework \cite{xu2024exploring} \cite{vishwamitra2024moderating}, we introduce a \textbf{multi-stage prompting strategy} that guides LLMs through a structured progression of understanding and decision-making regarding hateful content.

In the first stage, the LLM is primed with a clear role definition and task-specific context, which is critical for aligning its responses with the normative and cultural expectations of content moderation. The initial prompt, as shown below, defines the scope of hateful content and sets the LLM’s role:

\begin{tcolorbox}[colframe=green!70!black, colback=white, boxrule=1pt, arc=3pt]
You are a video moderation specialist tasked with assessing hateful content. The hateful content is defined as: Videos that incite discrimination, cause discomfort or distress, or demean individuals or groups based on attributes such as race, ethnicity, nationality, religion, disability, age, veteran status, sexual orientation, gender identity, etc.
\end{tcolorbox}

After grounding the model in this definition, we prompt it to analyze multimodal descriptions of the video scene, encouraging it to reason across visual, auditory, and textual modalities. The prompt is designed to leverage the LLM’s textual reasoning capabilities and is presented as follows:

\begin{tcolorbox}[colframe=green!70!black, colback=white, boxrule=1pt, arc=3pt]
Please combine the \textcolor{red}{\{video/frame/music/OCR\}} and \textcolor{red}{\{speech\}} descriptions, and analyze the visual, textual, and audio elements of the video to detect and flag any hateful content. Do not describe the literal content of the video; instead, focus on its implicit meanings and whether it conveys hateful expressions.
\end{tcolorbox}
The reasoning process is abstractly formulated as follows:
\begin{equation}
Rationale = \mathrm{MLLM}(v_i^T, v_i^F, v_i^{\mathrm{OCR}}, v_i^M, v_i^S)
\end{equation}
where $v_i^V$, $v_i^F$, $v_i^{\mathrm{OCR}}$, $v_i^M$, and $v_i^S$ denote the video-level, frame-level, OCR-derived, music, and speech captions, respectively. The output, denoted as \textit{Rationale}, captures the model’s intermediate reasoning grounded in multimodal cues.

Finally, the LLM is prompted to produce a hatefulness score based on the generated rationale. This scoring is formalized as:
\begin{equation}
s = \Phi_{\text{LLM}}(\text{Rationale}), \quad \in [0, 1],
\label{eq:stage_two}
\end{equation}
where $\Phi_{\text{LLM}}^{(2)}(\cdot)$ represents the scoring function implemented through LLM-based reasoning. The scoring prompt is shown below:
\begin{tcolorbox}[colframe=green!70!black, colback=white, boxrule=1pt, arc=3pt]
Given the \textcolor{red}{\{rationale\}}, if you were a content moderation specialist, how would you rate the scene on a scale from 0 to 1? Here, 0 denotes a standard (non-hateful) scene, 0.5 denotes an aggressive, offensive, or hostile scene, and 1 indicates an explicitly hateful scene.
\end{tcolorbox}

This multi-stage prompting design aims to improve the granularity, interpretability, and accuracy of hatefulness scoring in multimodal contexts by aligning model reasoning with task-specific definitions and modality-aware evidence.

\subsection{Frame-level Hateful Video Localization}

To achieve precise frame-level localization of hateful content, we first anchor our analysis around frame-level captions derived from the speech modality. For the remaining four captioning modalities (image, OCR, music, and video), we independently concatenate each modality-specific caption with the speech caption at each frame, constructing comprehensive multimodal contexts for scoring.

As mentioned in Multi-stage prompting, prior to scoring, we make the LLM summarize the concatenated captions. Once summarized multimodal captions are constructed, the LLM performs scoring independently on each modality using our multi-stage prompting approach. For every frame \( j \), this results in modality-specific hatefulness scores:
\begin{equation}
\begin{split}
    s_j^{m} &= \text{MultiStagePrompting}(C_j^{m, \text{sum}}), \\
    &\quad m \in \{\text{img}, \text{ocr}, \text{music}, \text{video}\}.
\end{split}
\end{equation}

After obtaining modality-specific hatefulness scores, we select the maximum score across all modalities to represent the hatefulness intensity for the current frame, formulated as:
\begin{equation}
    s_j^{\text{final}} = \max\{ s_j^{\text{img}}, s_j^{\text{ocr}}, s_j^{\text{music}}, s_j^{\text{video}} \}.
    \label{eq:max_fusion}
\end{equation} 

This strategy reflects a moderation principle: if any modality strongly indicates hatefulness, the frame should be flagged for review. By repeating this scoring process for each frame throughout the entire video, we obtain a complete temporal hatefulness profile. Finally, we apply a decision threshold \( \tau \) to binarize the frame-level predictions:
\begin{equation}
    H_j = \mathbb{I}(s_j^{\text{final}} > \tau),
    \label{eq:binary_localization}
\end{equation}
where \( \mathbb{I}(\cdot) \) is the indicator function. The value of \( \tau \) can be tuned based on operational constraints or validation performance. This yields a temporal sequence of hate annotations that can support video-level aggregation, visualization, or downstream policy enforcement.

\section{Experiments}

We validate our training-free method \textbf{LELA} on HateMM \cite{hatemm} and MultiHateClip(MHC) \cite{multihateclip} datasets, comparing its performance against state-of-the-art training-free baselines. To support our core design choices, particularly in the proposed multi-stage prompting and composition matching strategy, we conduct a comprehensive ablation study. Additionally, we perform modality-specific ablations and experiment with different LLMs for scoring, in order to assess the effectiveness of LLMs in multimodal hate video understanding.

% The remainder of this section is organized as follows: we first describe our experimental setup, including datasets and evaluation metrics. Section 4.1 presents and discusses the main results. Section 4.2 evaluate the performance on different backbone LLM models, and Section 4.3  presents the ablation analysis of our proposed framework. More details about the experiment could be found at appendix.

\begin{figure*}[t]
    \centering
    \includegraphics[width=\linewidth]{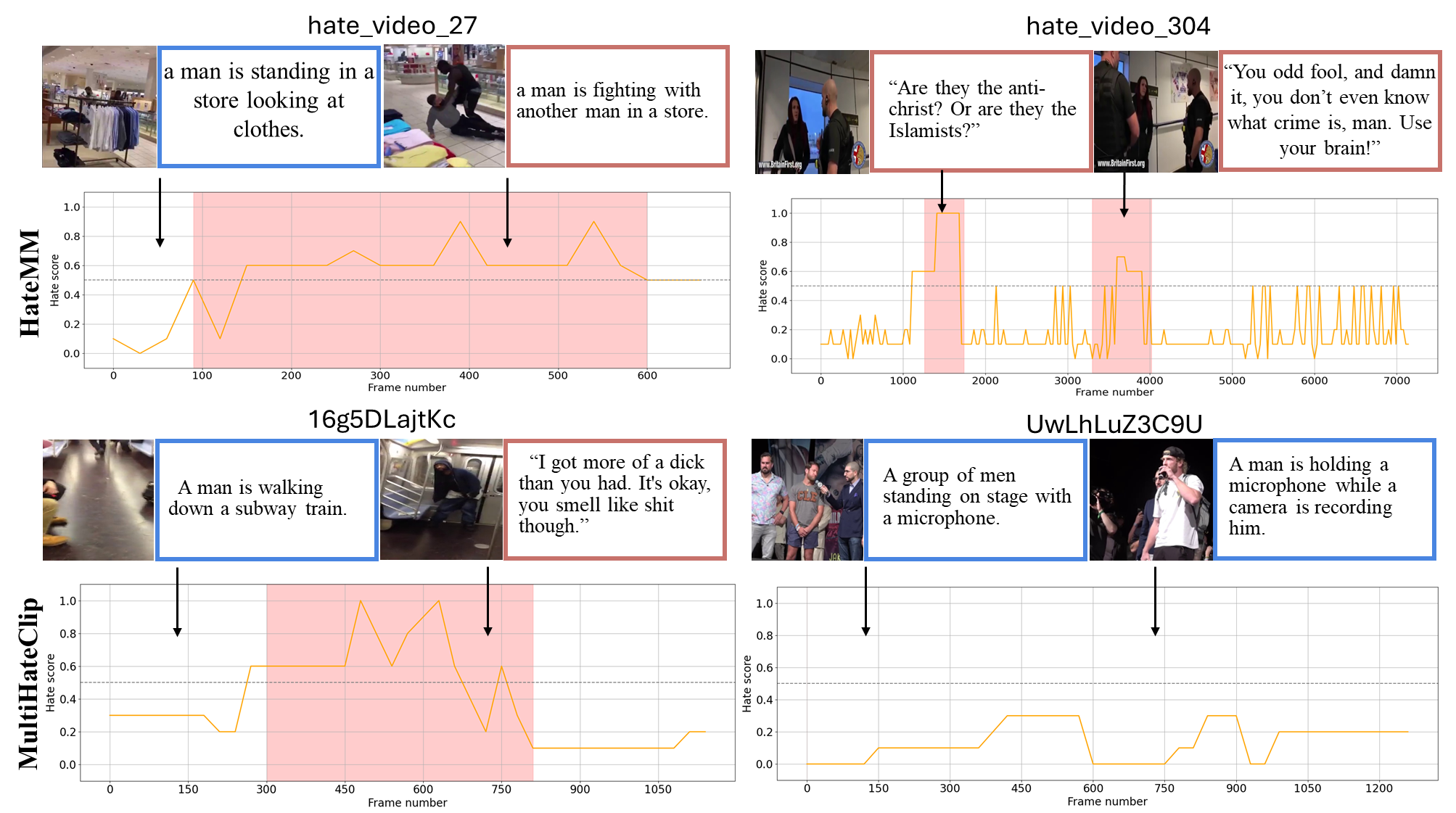}
    \caption{\small We showcase qualitative results obtained by our framework on four test videos, including two examples from the HateMM dataset (top row) and two from the MultiHateClip dataset (bottom row). For each video, we plot the predicted hate score across frames, as computed by our method. We display selected keyframes along with their most relevant modality caption. Frames predicted as non-hateful are marked with blue bounding boxes, while frames predicted as hateful are marked with red bounding boxes. Ground-truth hateful segments are also highlighted in pink, providing a reference to assess the accuracy and interpretability of our localization results.}
    \label{fig:quality}
\end{figure*}
\begin{table*}[t]
\centering
\renewcommand{\arraystretch}{1}
\resizebox{\linewidth}{!}{
\begin{tabular}{l c c c c c c c}
\toprule
\textbf{Method} & \textbf{PR-AUC (\%)} & \textbf{ROC-AUC (\%)} & \textbf{Acc} & \textbf{M-F1} & \textbf{F1(H)} & \textbf{P(H)} & \textbf{R(H)} \\
\midrule
\rowcolor{purple!10}
HTMM \cite{hatemm} & - & - & 0.7481 & 0.7353 & 0.6728 & 0.6954 & 0.6546 \\
\rowcolor{purple!10}
MHCL \cite{multihateclip} & - & - & 0.7503 & 0.7407 & 0.6238 & 0.6642 & 0.6831 \\
\rowcolor{purple!10}
HVGuard \cite{jing2025hvguard} & -& -&0.8563 & 0.8597 & 0.8479 & 0.8228 & 0.8009 \\
\rowcolor{purple!10}
Yue el al.\cite{yue2025multimodalhatedetectionusing}& -& -&0.821- & - & 0.771- & 0.798- & 0.754- \\
\rowcolor{purple!10}
CMFusion\cite{CMfusion}& -& -&0.823- & - & 0.860- & 0.817- & 0.908- \\
\rowcolor{purple!10}
Wang et al. \cite{wang2025cross} & -& -&0.82-- & 0.82-- & 0.80-- & 0.80-- & 0.79-- \\
\rowcolor{purple!10}
MM-HSD \cite{cespedes2025mm} & -& -&0.878- & 0.874- & 0.853- & 0.849- & 0.857- \\
\midrule
\rowcolor{green!10}
ZS-CLIP \cite{ZS-Clip} & 0.5216 & 0.5367 & 0.5019 & 0.6455 & 0.5699 & 0.5425 & 0.5215\\
\rowcolor{green!10}
ZS ImageBind \cite{girdhar2023imagebind} & 0.5237 & 0.5683 & 0.5317 & 0.4813 & 0.6419 & 0.5042 & 0.4981 \\
\rowcolor{green!10}
LLAVA-1.5 \cite{llava-1.5} & 0.5327 & 0.5529 & 0.5304 & 0.5149 & 0.4971 & 0.5653 & 0.4742 \\
\rowcolor{green!10}
LAVAD \cite{LAVAD} & 0.5781 & 0.6163 & 0.5716 & 0.5862 & 0.5319 & 0.5784 & \textbf{0.6827}\\
\rowcolor{green!10}
Lin el al. \cite{lin2025unified} & 0.6239 & 0.5674 & 0.6738 & 0.6127 & 0.5573 & 0.7132 & 0.4568\\
\rowcolor{green!10}
\textbf{LELA (ours)} & \textbf{0.7264} & \textbf{0.6756} & \textbf{0.7148} & \textbf{0.7043} & \textbf{0.6484} & \textbf{0.8152} & 0.5481 \\
\rowcolor{green!10}
\bottomrule
\end{tabular}
}
\caption{Comparison of supervised and zero-shot methods on HateMM dataset.}
\end{table*}
\begin{table*}[t]
\centering
\renewcommand{\arraystretch}{1}
\resizebox{\linewidth}{!}{
\begin{tabular}{l c c c c c c c}
\toprule
\rowcolor{white}
\textbf{Method} & \textbf{PR-AUC (\%)} & \textbf{ROC-AUC (\%)} & \textbf{Acc} & \textbf{M-F1} & \textbf{F1(H)} & \textbf{P(H)} & \textbf{R(H)} \\
\midrule
\rowcolor{purple!10}
HTMM \cite{hatemm} & - & -& 0.6861  & 0.7456 & 0.6817 & 0.6875 & 0.6597 \\
\rowcolor{purple!10}
MHCL \cite{multihateclip} & - & - & 0.7498 & 0.7317 & 0.6153 & 0.6750 & 0.6783 \\
\rowcolor{purple!10}
HVGuard \cite{jing2025hvguard} & -& -&0.8090 & 0.6646 & 0.4556 & 0.4722 & 0.5000 \\
\rowcolor{purple!10}
Yue el al.\cite{yue2025multimodalhatedetectionusing}& -& -&0.78-- & - & 0.77-- & 0.80-- & 0.77-- \\
\rowcolor{purple!10}
Wang et al. \cite{wang2025cross} & -& -&0.82-- & 0.81-- & 0.76-- & 0.87-- & 0.68-- \\
\midrule
\rowcolor{green!10}
ZS-CLIP \cite{ZS-Clip} & 0.5181 & 0.5449 & 0.5021 & 0.6395 & 0.5794 & 0.5371 & 0.5289 \\
\rowcolor{green!10}
ZS ImageBind \cite{girdhar2023imagebind} & 0.5135 & 0.5753 & 0.5391 & 0.4756 & 0.6392 & 0.5074 & 0.5014 \\
\rowcolor{green!10}
LLAVA-1.5\cite{llava-1.5} & 0.5319 & 0.5438 & 0.5342 & 0.5062 & 0.4974 & 0.5766 & 0.4694 \\
\rowcolor{green!10}
LAVAD \cite{LAVAD} & 0.5865 & 0.6302 & 0.5833 & 0.5799 & 0.5344 & 0.5671 & \textbf{0.6926} \\
\rowcolor{green!10}
Lin et al. \cite{lin2025unified} & 0.6147 & 0.5626 & 0.6174 & 0.5849 & 0.5637 & 0.7233 & 0.4813 \\

\rowcolor{green!10}
\textbf{LELA (ours)} & \textbf{0.7227} & \textbf{0.6733} & \textbf{0.7124} & \textbf{0.6923} & \textbf{0.6568} & \textbf{0.8217} & 0.5387 \\

\bottomrule
\end{tabular}
}
\caption{Comparison of supervised and zero-shot methods MHC dataset.}
\end{table*}
\subsection{Comparison with state of the art}

To the best of our knowledge, no existing method explicitly addresses the task of frame-level video hate localization.  To enable a fair and meaningful comparison with our training-free framework, we introduce a set of training-free baselines derived from vision-language models (VLMs), alongside state-of-the-art approaches in video anomaly detection (VAD). Additionally, we include supervised classification methods from the domain of hate video detection (HVD) for comprehensive evaluation.

\textbf{Performance Metrics.} We evaluate the performance of frame-level hate localization using two standard metrics: ROC-AUC and Average Precision (AP). ROC-AUC measures the area under the receiver operating characteristic curve computed at the frame level. It is a threshold-agnostic metric that evaluates the model’s ability to distinguish between hateful and non-hateful frames across all possible decision thresholds, making it particularly suitable for localization tasks. We also report \textbf{Average Precision (AP)}, defined as the area under the frame-level precision-recall curve. This metric is especially important in imbalanced scenarios where hateful content occurs sparsely. Our evaluation protocol follows the established practice introduced in~\cite{LAVAD}.

\textbf{Qualitative Results.} Fig.~\ref{fig:quality} presents qualitative results of our framework using sample videos from the HateMM and MultiHateClip datasets. We highlight keyframes along with their multimodal summaries to illustrate our model’s interpretability. In the hateful video examples (e.g., Row 1 and Row 2, Column 1), the frame-level summaries effectively capture the contextual cues of hostile, offensive, or aggressive content. These rich textual descriptions help the LLM accurately localize the hateful segments.  In contrast, for a non-hateful example, the summaries remain neutral, and our model consistently assigns low hate scores across all frames. This confirms the model’s capability to distinguish between harmful and benign content.

\textbf{Quantitative Results.} 
We evaluate the performance of both supervised and zero-shot methods on the HateMM and MHC datasets, presented in Table 1 and Table 2. Supervised baselines, such as HTMM and MHCL, demonstrate strong accuracy and F1 scores but fall short in PR-AUC performance. In contrast, our proposed \textbf{LELA} framework consistently outperforms other zero-shot models across all metrics and even achieves results comparable to supervised approaches. These findings highlight LELA's capability to effectively capture and integrate multimodal signals across diverse modalities.

\begin{table*}[t]
\centering
\caption{Performance comparison of LLMs on HateMM dataset.}
\resizebox{\linewidth}{!}{
\begin{tabular}{l|c|ccccccc}
\hline
Methods& LLM Size & ROC\_AUC & PR\_AUC & Acc & M-F1 & F1(H) & P(H) & R(H) \\
\hline
\multicolumn{9}{c}{Open-Source LLMs}\\
DeepSeek-R1-1.5B \cite{deepseek}     & 1.5B & 55.69 & 49.60 & 56.99 & 54.90 & 45.21 & 55.03 & 38.37 \\
DeepSeek-R1-7B  \cite{deepseek}      & 7B   & 64.73 & 60.58 & 64.67 & 63.37 & 56.67 & 71.33 & 48.89 \\
Qwen2.5-3B \cite{qwen}               & 3B   & 57.49 & 51.42 & 58.60 & 56.70 & 47.62 & 58.33 & 40.23 \\
Qwen2.5-7B \cite{qwen}               & 7B   & 62.14 & 57.29 & 63.98 & 62.49 & 55.03 & 68.33 & 46.07 \\
LLaMA-2 7B \cite{llama}              & 7B   & 60.97 & 54.23 & 61.29 & 59.60 & 51.35 & 63.33 & 43.18 \\
\hline
\multicolumn{9}{c}{Proprietary LLMs} \\
Gemini-2.0 Flash \cite{gemini}       & -    & 70.28 & \textbf{68.98} & 70.97 & 70.08 & 64.94 & \textbf{83.33} & 53.19 \\
\textbf{GPT-4o Mini \cite{gpt}}      & - & \textbf{72.64} & 67.56 & \textbf{71.51} & \textbf{70.46} & \textbf{64.90} & 81.67 & \textbf{54.85} \\
\hline
\end{tabular}
}
\label{tab:hatemm_performance}
\end{table*}

\begin{table*}[t]
\centering
\caption{Performance comparison of LLMs on MHC dataset.}
\resizebox{\linewidth}{!}{
\begin{tabular}{l|c|ccccccc}
\hline
Methods & LLM Size & ROC\_AUC & PR\_AUC & Acc & M-F1 & F1(H) & P(H) & R(H)\\
\hline
\multicolumn{9}{c}{Open-Source LLMs} \\
DeepSeek-R1-1.5B \cite{deepseek}     & 1.5B & 55.41 & 49.18 & 56.62 & 54.47 & 45.20 & 55.07 & 38.12 \\
DeepSeek-R1-7B  \cite{deepseek}      & 7B   & 63.28 & 59.33 & 63.70 & 62.41 & 55.92 & 70.81 & 48.05 \\
Qwen2.5-3B \cite{qwen}               & 3B   & 57.03 & 51.37 & 58.21 & 56.35 & 47.43 & 58.26 & 40.02 \\
Qwen2.5-7B \cite{qwen}               & 7B   & 61.82 & 56.51 & 62.41 & 61.21 & 53.77 & 67.45 & 45.12 \\
LLaMA-2 7B \cite{llama}              & 7B   & 61.42 & 53.97 & 60.92 & 59.27 & 50.89 & 63.02 & 42.87 \\
\hline
\multicolumn{9}{c}{Proprietary LLMs} \\
Gemini 1.5 Pro \cite{gemini}         & -    & 70.57 & 67.12 & 70.84 & 70.01 & \textbf{65.82} & \textbf{83.01} & 53.05 \\
\textbf{GPT-4o Mini \cite{gpt}}      & \textbf{-} & \textbf{72.27} & \textbf{67.33} & \textbf{71.48} & \textbf{70.43} & 64.84 & 81.52 & \textbf{54.81} \\
\hline
\end{tabular}
}
\label{tab:MHC_performance}
\end{table*}
\begin{table*}[t]
\centering
\renewcommand{\arraystretch}{1}
\resizebox{0.5\linewidth}{!}{
\begin{tabular}{cc}
\hline
Decision Threshold & Accuracy \\
\hline
0.3 & 0.4287 \\
0.4 & 0.5153 \\
0.5 & 0.7148 \\
0.6 & 0.6789 \\
0.7 & 0.5482 \\
\hline
\end{tabular}
}
\caption{Accuracy under different decision thresholds.}
\end{table*}

\textbf{\begin{table}[ht]
\centering
\renewcommand{\arraystretch}{1}
\resizebox{\linewidth}{!}{  % <-- Scale table width to 95%
\begin{tabular}{cccc|c}
\toprule
\textbf{Contextualization} & \textbf{Rationale Generation} & \textbf{final decision} &  & \textbf{AUC(\%)} \\
\midrule
\textcolor{red}{\ding{55}} & \textcolor{red}{\ding{55}} & \textcolor{green}{\checkmark} &  & 56.48 \\
\textcolor{green}{\checkmark} & \textcolor{red}{\ding{55}} & \textcolor{green}{\checkmark} &  & 65.82 \\
\textcolor{red}{\ding{55}} & \textcolor{green}{\checkmark} & \textcolor{green}{\checkmark}  &  & 59.70 \\
\textcolor{green}{\checkmark} & \textcolor{green}{\checkmark} & \textcolor{green}{\checkmark} &  & \textbf{68.28} \\
\bottomrule
\end{tabular}
}
\caption{Ablation study on the multi-stage prompting strategy. Row 4 represents the full model. AUC refers to ROC\_AUC.}
\label{tab:multistage}
\end{table}}

\begin{table}[b]
\centering
\caption{Ablation study on modality composition using GPT-4o Mini as the backbone LLM.}
\label{tab:composition_matching}
\resizebox{0.5\linewidth}{!}{
\begin{tabular}{l|c}
\hline
Modality Composition & AUC (\%) \\
\hline
Speech (baseline)      & 68.28 \\
+ Image                & 68.89 \\
+ OCR                  & 71.47 \\
+ Music                & 71.75 \\
+ Video                & \textbf{72.27} \\
\hline
\end{tabular}
}
\end{table}

\subsection{Evaluation of different LLMs}

We evaluate the hate understanding capabilities of LLMs on two benchmarks: HateMM and MHC English (Tables~\ref{tab:hatemm_performance} and \ref{tab:MHC_performance}). On HateMM, \textbf{DeepSeek-R1-7B} leads among open-source models, achieving 64.73\% ROC\_AUC, 60.58\% PR\_AUC, 64.67\% accuracy, and a macro-F1 of 63.37\%, with strong precision (71.33\%) but moderate recall (48.89\%). \textbf{Qwen2.5-7B} shows comparable performance, while smaller variants like \textbf{DeepSeek-R1-1.5B} and \textbf{Qwen2.5-3B} perform significantly worse. Similar trends hold on the MHC English dataset, where \textbf{DeepSeek-R1-7B} again outperforms other open-source models (63.28\% ROC\_AUC, 59.33\% PR\_AUC), followed by \textbf{Qwen2.5-7B}, while smaller models continue to struggle.

Across both datasets, \textbf{proprietary LLMs} demonstrate a clear advantage. \textbf{GPT-4o Mini} achieves the best results overall (e.g., 72.64\% ROC\_AUC on HateMM, 72.27\% on MHC), with high precision (81.67\%, 81.52\%) and balanced F1(H) scores (64.90\%, 64.84\%). \textbf{Gemini-2.0 Flash} also shows strong performance, particularly in precision, but slightly lags in recall and consistency. Based on the analysis, we select GPT-4o Mini as the backbone LLM in our experimental setting.

\subsection{Ablation study}

\textbf{Multi-stage prompting.}
To validate the effectiveness of our proposed multi-stage prompting mechanism, we conduct a systematic ablation study by decomposing the overall process into three functional components: Contextualization, Rationale Generation, and Final Decision. Table~\ref{tab:multistage} presents the performance of four prompting strategies by selectively enabling or disabling individual components. When only the final decision is performed without prior context or reasoning, the model yields the lowest AUC of 56.48\%, indicating that direct scoring is insufficient for reliably assessing hateful content. Enabling contextualization alongside the final decision results in a substantial improvement, with the AUC increasing to 65.82\%. Additionally, using rationale generation in isolation improves performance to 59.70\%, indicating that intermediate reasoning is beneficial. The full model with all three components achieves the highest AUC of 68.28\%, demonstrating the complementary nature of contextual grounding and intermediate reasoning. These findings collectively highlight that both contextual guidance and structured reasoning are essential for accurate and robust hatefulness scoring in multimodal scenarios.

\textbf{Composition Matching.} We further investigate the effectiveness and necessity of each modality component in our proposed multimodal framework through composition matching experiments. Specifically, starting from baseline speech captions analyzed by our multi-stage prompting mechanism, we incrementally incorporate additional modalities: visual (image frames), OCR-extracted text, background music, and finally, full video-level context. All experiments utilize GPT-4o Mini as the backbone LLM, isolating the contribution of modality composition to the final performance. Table~\ref{tab:composition_matching} summarizes these results.
\subsection{Sensitivity analysis}
For LLM score decisions, we set 0.5 as the threshold for both qualitative and quantitative experiments. To evaluate the robustness of our model under different decision boundaries, we conduct a sensitivity analysis by measuring classification accuracy across a range of threshold values. As shown in Table 5, accuracy peaks at the standard threshold of 0.5, with notable declines observed as the threshold shifts lower or higher. This indicates that the model performs most reliably at the 0.5 threshold while exhibiting moderate sensitivity to threshold variation, suggesting that the LLM has effectively established a coherent scalar scoring system.

\section{Conclusion}

In this paper, we propose \textbf{LELA}, a novel training-free framework for frame-level hateful content localization in videos. LELA leverages multimodal caption decomposition and multi-stage prompting to elicit structured reasoning from large language models. Extensive evaluations on HateMM and MultiHateClip datasets demonstrate that LELA outperforms. This work underscores the potential of language model-based reasoning for complex content moderation tasks and paves the way for future research in scalable, training-free multimodal understanding.

%Bibliography
\bibliographystyle{unsrt}  
\bibliography{reference}

\end{document}